\documentclass[10pt,twocolumn,letterpaper]{article}

\usepackage{cvpr}              

\usepackage{graphicx}
\usepackage{xcolor}
\usepackage{amsmath}
\usepackage{amssymb}
\usepackage{caption}
\usepackage{floatrow}
\usepackage{enumitem}
\usepackage{pifont}
\usepackage{float}
\floatstyle{plaintop}
\restylefloat{table}
\usepackage{booktabs}
\usepackage{txfonts} 
\usepackage{xurl}

\newcommand{\edit}[1]{{#1}}
\newcommand{\cmark}{\ding{51}}%
\newcommand{\xmark}{\ding{55}}%
\newcommand{\boldstart}[1]{\noindent\textbf{#1}}

\newcommand{\neck}{\kappa}
\newcommand{\jaw}{\gamma}
\newcommand{\expr}{\xi}
\newcommand{\Gp}{G_p}
\newcommand{\Mp}{W_p}

\newcommand{\pitch}{\phi}
\newcommand{\yaw}{\psi}

\newcommand{\rot}{\alpha^\mathrm{rot}}

\newcommand{\wrot}{w_\mathrm{rot}}
\newcommand{\wexpr}{w_\mathrm{expr}}
\newcommand{\rotmlp}{\mathcal{F}_\mathrm{rot}}
\newcommand{\exprmlp}{\mathcal{F}_\mathrm{expr}}

\newcommand{\wfinal}{w_\textrm{final}}
\newcommand{\wplus}{w^+}

\newcommand{\exprloss}{\mathcal{L}_\mathrm{expr}}
\newcommand{\poseloss}{\mathcal{L}_\mathrm{pose}}
\newcommand{\localloss}{\mathcal{L}_\mathrm{local}}
\newcommand{\consistencyloss}{\mathcal{L}_\mathrm{cons}}

\newcommand{\wpose}{\lambda_\mathrm{pose}}
\newcommand{\wconsistency}{\lambda_\mathrm{cons}}
\newcommand{\wexprloss}{\lambda_\mathrm{expr}}
\newcommand{\wlocal}{\lambda_\mathrm{local}}
\newcommand{\lpips}{\mathcal{L}_\mathrm{LPIPS}}
\newcommand{\pixel}{\mathcal{L}_{L2}}
\newcommand{\mse}{\mathcal{L}_{L2}}
\newcommand{\wptimse}{\lambda^\mathrm{P}_{L2}}

\newcommand{\ptireg}{\mathcal{L}_R}
\newcommand{\wptireg}{\lambda^\mathrm{P}_R}
\newcommand{\idloss}{\mathcal{L}_\mathrm{id}}
\newcommand{\wlpips}{\lambda_\mathrm{LPIPS}}
\newcommand{\wpixel}{\lambda_{L2}}

\newcommand{\widloss}{\lambda_\mathrm{id}}
\newcommand{\im}{I}
\newcommand{\imout}{I'}

\newcommand{\minus}{\scalebox{0.75}[1.0]{$-$}}

\DeclareMathOperator*{\argmin}{arg\,min}
\newcommand{\forapproval}[1]{}
\usepackage[pagebackref,breaklinks,colorlinks]{hyperref}

\usepackage[capitalize]{cleveref}
\crefname{section}{Sec.}{Secs.}
\Crefname{section}{Section}{Sections}
\Crefname{table}{Table}{Tables}
\crefname{table}{Tab.}{Tabs.}
\Crefname{figure}{\textbf{Fig.}}{\textbf{Figs.}}

\def\cvprPaperID{*****} 
\def\confName{CVPR}
\def\confYear{2023}

\begin{document}

\title{PVP: Personalized Video Prior for \quad \\ Editable Dynamic Portraits using StyleGAN}


\author
{
Kai-En Lin$^{1}$
\and
Alex Trevithick$^{1}$
\and
Keli Cheng$^{2}$
\and
Michel Sarkis$^{2}$
\and
Mohsen Ghafoorian$^{2}$
\and
Ning Bi$^{2}$
\and
Gerhard Reitmayr$^{2}$
\and
Ravi Ramamoorthi$^{1}$
\and
$^1$University of California, San Diego \quad \quad $^2$Qualcomm Technologies Inc.
}


\twocolumn[{%
\renewcommand\twocolumn[1][]{#1}%
\maketitle
\begin{center}
    \centering
    \captionsetup{type=figure}
     \includegraphics[width=\linewidth]{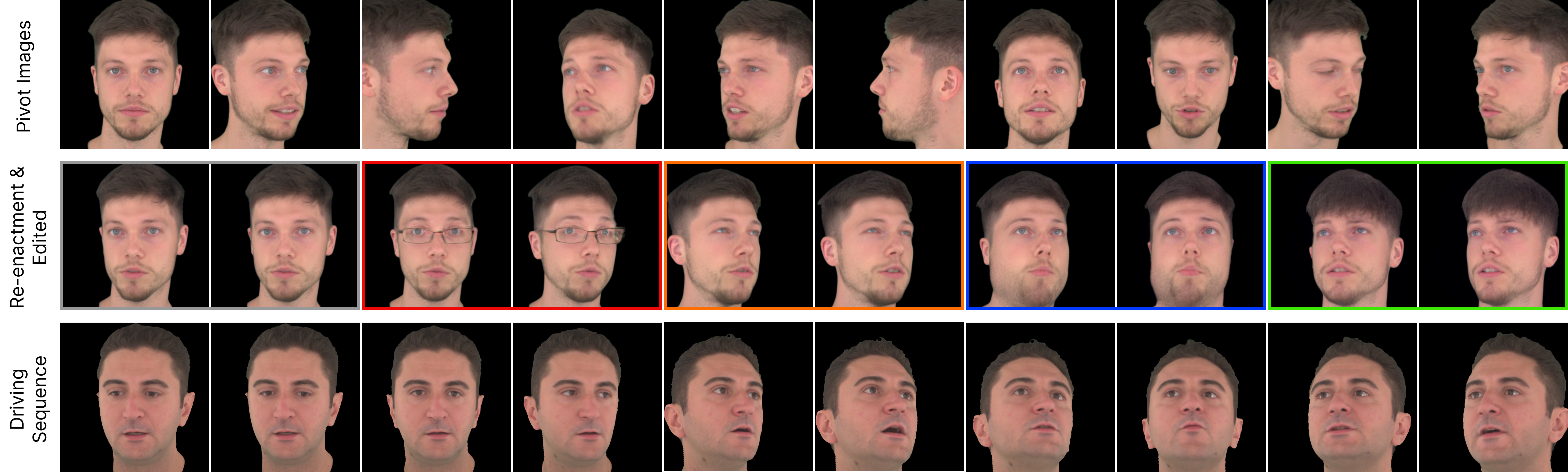}
      \caption{
      \boldstart{Portrait reenactment and editing with our proposed method.}
      Our method takes a monocular video and learns a personalized video prior, which allows accurate reeanctment and editing through StyleGAN.
      In the first row, we show the selected pivots used to fine-tune the StyleGAN generator.
      We show different editing results in the second row: \textcolor{gray}{original}, \textcolor{red}{eye glasses}, \textcolor{orange}{small eyes}, \textcolor{blue}{chubby}, \textcolor{green}{fringe hair}.
      The reenacted sequence is driven by the pose and expression of the third row.
      }
      \label{fig:teaser}
\end{center}%
}]
\maketitle

\begin{abstract}
    Portrait synthesis creates realistic digital avatars which enable users to interact with others in a compelling way.
    Recent advances in StyleGAN and its extensions have shown promising results in synthesizing photorealistic and accurate reconstruction of human faces.
    However, previous methods often focus on frontal face synthesis and most methods are not able to handle large head rotations due to the training data distribution of StyleGAN.
    In this work, our goal is to take as input a monocular video of a face, and create an editable dynamic portrait able to handle extreme head poses.
    The user can create novel viewpoints, edit the appearance, and animate the face.  
    Our method utilizes pivotal tuning inversion (PTI) to learn a personalized video prior from a monocular video sequence.
    Then we can input pose and expression coefficients to MLPs and manipulate the latent vectors to synthesize different viewpoints and expressions of the subject.
    We also propose novel loss functions to further disentangle pose and expression in the latent space.
    Our algorithm shows much better performance over previous approaches on monocular video datasets, and it is also capable of running in real-time at 54 FPS on an RTX 3080.
 
\end{abstract}  
\section{Introduction}

\begin{table*}[t]
  \centering
  \resizebox{1.0\textwidth}{!}{%
  \begin{tabular}{l l c c c c c}
  \toprule
    Category & Methods & \edit{Head pose editing} & Appearance editing & Monocular input & Riggable & Personalization \\
    \midrule
    NeRF & HeadNeRF~\cite{hong2021headnerf}     & \xmark & \cmark & \xmark & \xmark & \xmark \\
    & RigNeRF~\cite{athar2022rignerf}      & \cmark & \xmark & \cmark & \cmark & \cmark \\
    & FDNeRF~\cite{zhang2022fdnerf}        & \cmark & \xmark & \cmark & \cmark & \xmark \\
    & cGOF~\cite{sun2022cgof}              & \xmark & \cmark & \cmark & \cmark & \xmark \\
    & NeRFBlendShape~\cite{Gao2022nerfblendshape} & \cmark & \xmark & \cmark & \cmark & \cmark \\
    & NeRFFaceEditing~\cite{jiang2022nerffaceediting} & \xmark & \cmark & \cmark & \xmark & \cmark \\
    Parametric Model & NHA~\cite{grassal2021neural}         & \cmark & \xmark & \cmark & \cmark & \cmark \\
    & ROME~\cite{Khakhulin2022ROME}        & \xmark & \xmark & \cmark & \cmark & \xmark \\
    StyleGAN & FreeStyleGAN~\cite{FreeStyleGAN2021} & \xmark & \cmark & \xmark & \xmark & \xmark \\
    & PIRenderer~\cite{ren2021pirenderer}  & \cmark & \xmark & \cmark & \cmark & \xmark \\
    & LIA~\cite{wang2022latent}            & \cmark & \xmark & \cmark & \cmark & \xmark \\
    & StyleHEAT~\cite{styleheat}           & \cmark & \cmark & \cmark & \cmark & \xmark \\
    & \textbf{Ours}                        & \cmark & \cmark & \cmark & \cmark & \cmark \\
    \bottomrule
  \end{tabular}
  }
  \caption{\boldstart{Comparison of different methods.}
  Our proposed method bridges the gap between 2D and 3D methods by enabling reconstruction on extreme viewpoints through personalization.
  2D methods allow for efficient image synthesis from single view inputs, but they often cannot handle larger motions.
  Even if the input video contains large motion changes, 2D methods~\cite{wang2022latent,styleheat} are not able to use the information and reconstruct more difficult viewpoints.
  On the other hand, 3D methods~\cite{grassal2021neural,Gao2022nerfblendshape} can handle difficult viewpoints and NeRFBlendShape can achieve some form of personalization in expression.
  However, they do not allow full editability on appearance like the 2D methods.
  Our method achieves editable head poses, appearance editing via StyleGAN, monocular input, riggability through FLAME-like controls, and personalization with the help of a personalized video prior.
  }\label{tab:comparison}
\end{table*}



Digital avatars offer a compelling way to represent human appearance and expression, enabling various applications, like telepresence and virtual reality.
Recent advances in StyleGAN~\cite{karras2019style,Karras2019stylegan2} and neural radiance fields (NeRFs)~\cite{nerf} have accelerated the development of digital avatars~\cite{hong2021headnerf,athar2022rignerf}.
Radiance fields have also spawned many extensions~\cite{hong2021headnerf,athar2022rignerf,zhang2022fdnerf,Gao2022nerfblendshape,xu2022manvatar} which capture various characteristics of human portraits, such as expression, appearance and identity.

However, the NeRF-based methods mostly focus on reconstructing the input sequence and do not provide a good way to edit or manipulate various properties. Many of these approaches rely on learning a 3D generative prior, thus requiring inversion into the latent space, which requires generator finetuning and an inevitable loss of 3D representation ability. These methods are also non-trivial to extend to video sequences of a subject.
We argue that editability also plays an important part in making the digital avatars engaging, and it offers the user more control over their desired looks (see Fig.~\ref{fig:teaser}).
A recent paper, FreeStyleGAN~\cite{FreeStyleGAN2021}, enables editing on static human portraits, but it requires multiple carefully-shot images with the subject holding still for several seconds. 
Concurrent work~\cite{sun2022nerfeditor,jiang2022nerffaceediting} provides an editing interface to allow user inputs, but these methods are not animatable (e.g. enabling 3DMM-like control~\cite{blanz1999morphable}) and thus do not offer the same level of interactivity.
Lastly, while ClipFace~\cite{aneja2022clipface} provides editability on 3D morphable models (3DMM) and their texture, it does not handle other facial features like hair.

Our proposed method offers a new way to tackle portrait synthesis using personalized StyleGAN from monocular portrait videos.
First, we carefully sample several frames from the input videos to use as pivots, and the pivots are used to perform PTI~\cite{roich2021pivotal} and fine-tune the StyleGAN generator to produce a personalized manifold, allowing for faithful reconstruction of the subject under extreme poses and smooth transitions between different head poses.
Then, we employ lightweight pose and expression encoders to enable finer control over the representation.
We can render novel head poses and expressions in real-time (at 54 FPS) by feeding in the corresponding pitch, yaw angles and FLAME coefficients~\cite{FLAME:SiggraphAsia2017}.
We also propose a novel expression matching loss and pose consistency loss to disentangle the editing directions in the latent space, making it easier to change one attribute at a time.
Additionally, exploiting various methods in StyleGAN editing~\cite{patashnik2021styleclip,shen2020interfacegan}, our method can provide good editing capability.
Project website and code: \url{https://cseweb.ucsd.edu/~viscomp/projects/EGSR23PVP/}

We summarize our contributions as follows:
\begin{enumerate}
    \item a personalized video prior derived from monocular portrait video of a given subject (Sec.~\ref{sec:manifold});
    \item a novel algorithm that enables control of a dynamic portrait within the personalized manifold, allowing editing on pose, expression and appearance (Sec.~\ref{sec:mapper}, Fig.~\ref{fig:pipeline});
    \item expression matching and pose consistency loss to better disentangle the poses and expressions given only a short portrait video of a subject (Sec.~\ref{sec:loss}).
\end{enumerate}

\section{Related Work}

Previous methods have shown promising results in reconstructing photorealistic face renderings.
In particular, there exist 4 promising directions:
(a) StyleGAN models which can synthesize 2D face renderings and allow for user control by traversing in their latent space (Sec.~\ref{sec:stylegan2d});
(b) neural radiance fields (NeRF) that handle the dynamic facial expression and complex visual effects through volumetric rendering (Sec.~\ref{sec:nerf});
(c) parametric models, like 3DMM and FLAME, which provide explicit pose and expression control through skinning (Sec.~\ref{sec:parametric_face_models}).
\edit{(d) facial reenactment methods which enable facial expressions to be transferred to different subjects (Sec.~\ref{sec:facial_reenactment}).}
We give an overview of these methods in the following subsections and a comparison of our method against previous methods in Table~\ref{tab:comparison}.


\subsection{Face Synthesis with StyleGAN}~\label{sec:stylegan2d}

Generative models like StyleGAN2~\cite{Karras2019stylegan2} have demonstrated an impressive capability of synthesizing photorealistic portraits while only trained on in-the-wild imagery.
In order to reconstruct human faces, a common method is to invert the underlying latent code.
GAN inversion methods, like e4e~\cite{tov2021designing} and pSp~\cite{richardson2021encoding}, have shown remarkable results in finding the most representative latent code.
With the latent codes, PTI~\cite{roich2021pivotal} and MyStyle~\cite{nitzan2022mystyle} further optimize the StyleGAN generator to learn a personalized latent space, enabling editing while staying faithful to the same identity.
In terms of controlling the pose and expression of the StyleGAN renderings, StyleRig~\cite{tewari2020stylerig} predicts modified latent codes directly from semantic control parameters.
StyleHEAT~\cite{styleheat} performs warping on the intermediate features layers of the StyleGAN synthesis network.
Although these methods show promising results, there are some shortcomings.
For example, StyleRig could introduce a shift in identity after rotating the head pose.
The warping scheme used by StyleHEAT could not handle large viewpoint changes well enough (see Fig.~\ref{fig:qualitative}).
The aforementioned StyleGAN methods are mainly handling 2D images.

While 3D GAN-based methods~\cite{piGAN2021,Chan2021,gu2021stylenerf,orel2022styleSDF,deng2022gram} generate photorealistic 3D representations from 2D images, \edit{the editability of such models requires further research to reach its full potential.} 
SofGAN~\cite{sofgan}, IDE-3D~\cite{ide3d}, FENeRF~\cite{sun2022fenerf} and CIPS-3D~\cite{zhou2021cips3d} demonstrated some level of editability on the appearance and expression of 3D GANs.
Additionally, inverting a video into their latent space is non-trivial as each frame is inverted independently, and 3D GAN PTI often collapses to a 2D representation.

As a result, our paper looks at lifting 2D StyleGAN renderings to a 3D representation that allows editing of both the facial appearance and expression.
One related work is FreeStyleGAN~\cite{FreeStyleGAN2021}.
However, it requires the subject to be static and multi-view input images.
Our method seeks to handle dynamic portraits with only a single-view input video.
Please refer to Table~\ref{tab:comparison} for a comparison of different methods.

\subsection{3D Methods for Dynamic Portraits}~\label{sec:nerf}

NeRF~\cite{nerf} brought about exciting advancements in the field of image-based rendering.
Recently, a plethora of work tackling portrait reconstruction~\cite{hong2021headnerf,athar2022rignerf,zhang2022fdnerf,Gao2022nerfblendshape,sun2021nelf,zheng2022imavatar,park2021nerfies,park2021hypernerf} has emerged.
First, NeRFace~\cite{Gafni_2021_CVPR} conditions the MLP with learnable codes and expression coefficients to represent dynamic facial motions.
HeadNeRF~\cite{hong2021headnerf} encodes identity, expression, albedo and illumination into latent codes and uses them to condition the NeRF MLP, providing some level of user control.
RigNeRF~\cite{athar2022rignerf} combines the 3DMM deformation fields with NeRF to allow explicit control of the expression and head poses.
Nerfies~\cite{park2021nerfies} and HyperNeRF~\cite{park2021hypernerf} provide ways to encode deformable NeRFs, enabling capture of 3D facial movements. 
Following the above methods, FDNeRF~\cite{zhang2022fdnerf} further extends to few-shot inputs and NeRFBlendShape~\cite{Gao2022nerfblendshape} blends hashgrids with different expression coefficients to allow for fast and expressive portrait reconstruction.
While the above approaches mostly focus on re-enactment and pose manipulation, they do not provide comprehensive appearance editing.

On the other hand, NeRFFaceEditing~\cite{jiang2022nerffaceediting} focuses on the editability of different facial parts.
SofGAN~\cite{sofgan} provides finer control over different facial parts, including appearance, shape and lighting effects.
However, these methods do not offer explicit facial pose control.
Our work aims to achieve granular pose control and appearance editing at the same time.


\subsection{Parametric Model for Human Faces}~\label{sec:parametric_face_models}

There has been a lot of interest in using morphable face models  to enable facial animation~\cite{tran2018nonlinear,garrido2014automatic,weise2011realtime}.
The FLAME model~\cite{FLAME:SiggraphAsia2017} offers a good trade-off between controllability and expressiveness.
Derivatives like DECA~\cite{DECA:Siggraph2021} can infer FLAME parameters given only a single image.
ROME~\cite{Khakhulin2022ROME} further expands DECA to include displacements for the head mesh and preserve more details in non-facial regions (e.g. hair, neck and shoulders).
A notable work in this category is NHA~\cite{grassal2021neural}, which optimizes a head model given a single-view video.
NHA provides a good geometry estimate of the subject while keeping explicit control over the FLAME parameters.
Although NHA supports editing the facial expression and head pose through the FLAME parameters, it does not have an interface for appearance editing.
Our proposed method focuses on this aspect and seeks to combine the editability of StyleGAN images with FLAME-like control parameters.

\subsection{Facial Reenactment Techniques}~\label{sec:facial_reenactment}
\edit{
In addition to the above methods, there are some other techniques that focus on the facial reenactment tasks.
In other words, these methods transfer the expressions from one subject to another.
Face2Face~\cite{thies2016face} achieves real-time face capture and reenactment by fitting a parametric 3D model to monocular RGB input videos.
Deep Video Portraits~\cite{kim2018deep} utilizes a translation network to convert coarse facial renderings into photorealistic video portraits.
Different from these methods, our work enables reenactment and editing simultaneously through the use of pose and expression encoder networks and the StyleGAN latent space.
}


\section{Background}\label{sec:background}

\begin{figure*}[t]
\centering
\includegraphics[width=1.0\textwidth]{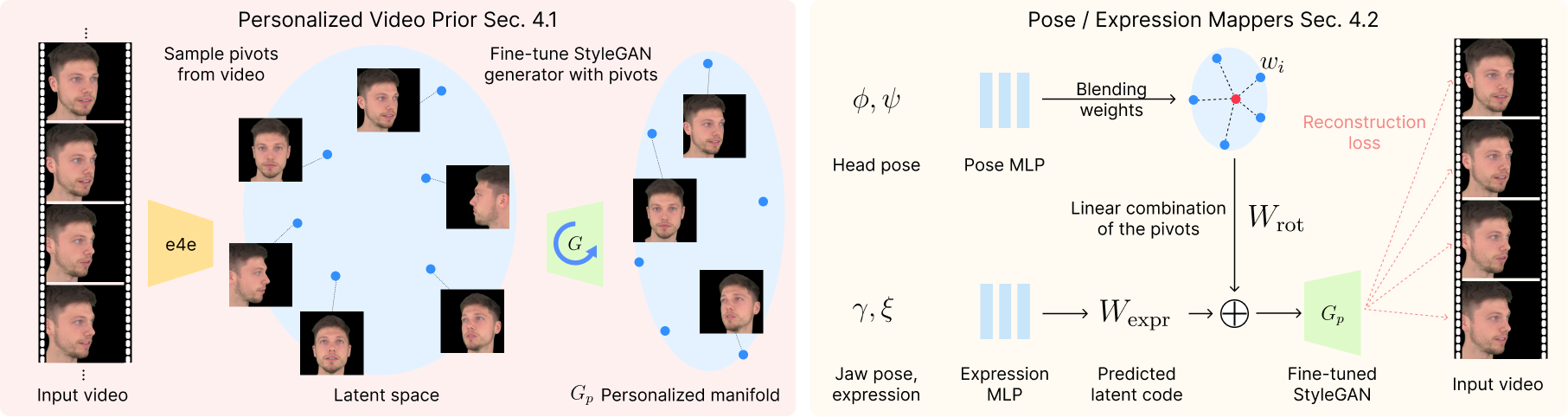}
\caption{\boldstart{Overview of our proposed method.}
Our algorithm is divided into two stages.
First, we use PTI to fine-tune the StyleGAN generator with the selected frames from the input video (see Sec.~\ref{sec:manifold}).
The fine-tuned StyleGAN now has a personalized manifold that reconstructs the pivots faithfully, and we utilize its smoothness to interpolate between the pivots and represent different head poses.
Then, we train the pose MLP to output the blending weights of the personalized manifold to provide a latent code that represents the rotated head.
We use the expression MLP to calculate the $W+$ latent code residuals (see Sec.~\ref{sec:mapper}).
Finally, the combined latent code is sent to the fine-tuned StyleGAN to synthesize the final rendering, and we supervise it with reconstruction loss and other supervision (see Sec.~\ref{sec:loss}).
}
\label{fig:pipeline}
\end{figure*}

In this section, we give a brief overview on StyleGAN and its latent space design.
Then, as we seek to edit real portraits, we introduce pivotal tuning inversion (PTI), which is the state-of-the-art method to perform GAN inversion, while keeping good editability.
PTI serves as an important foundation for our method, since we would like to explore the idea of a personalized video prior by fine-tuning the StyleGAN generator on selected frames of a given video (more in Sec.~\ref{sec:algorithm}).
Lastly, we briefly go through editing on the StyleGAN.

StyleGAN~\cite{Karras2019stylegan2} demonstrates high-quality image synthesis results, and it has a well-structured latent space, which allows smooth transition between different latent codes through linear interpolation.
To be more specific, it takes a latent code $z\in \mathbb{R}^{512}$ as input (often referred to as $\mathcal{Z}$ space) to a mapper network, which maps the input to the intermediate latent code $w\in \mathcal{W}$. 
The latent code $w$ is then affine transformed and fed to each convolutional layer in the synthesis network via adaptive instance normalization (AdaIN)~\cite{huang2017adain}.
The affine transformed latent code is often referred to as the $\mathcal{W}+$ space. We can then write the StyleGAN image synthesis as $I' = G(w^+(z); \theta)$, for a StyleGAN generator $G$ with weights $\theta$.

A way to adapt StyleGAN to editing real images is through GAN inversion.
The idea is to freeze the StyleGAN weights and optimize for the latent code in either $\mathcal{W}$~\cite{Karras2019stylegan2}, or $\mathcal{W}+$~\cite{abdal2019image2stylegan}.
However, there is a tradeoff between the two methods.
$\mathcal{W}$ space inversion offers better editability but poor reconstruction quality, whereas $\mathcal{W+}$ provides superior reconstruction, yet produces inferior editing results.
To address this tradeoff, the better way is to find pivot latent codes in the $\mathcal{W}+$ space via e4e~\cite{tov2021designing} and then fine-tune the StyleGAN generator based on the pivots, namely pivotal tuning inversion.
With PTI, it is possible to extend StyleGAN to handle samples which differ greatly from the training distribution, for instance, persons with makeup and faces viewed at large angles ($\ge 60^\circ$).
To be more specific, we can define the pivots as
\begin{equation}
    \wplus = E(\im),
\end{equation}
where $E$ is the e4e encoder, $\im$ the input image, and $\wplus$ the inverted latent code in $\mathcal{W}+$ space.
And we can synthesize the rendering with the StyleGAN $G$ by
\begin{equation}
    I' = G(\wplus;\theta),
\end{equation}
where $\theta$ denotes the weights of a StyleGAN generator.
Once we acquire the pivots, we can then optimize the StyleGAN generator $G$ and obtain the fine-tuned weights $\theta_p$ with the following objective:
\begin{equation}\label{eq:pti_obj}
    \theta_p = \argmin_\theta \lpips(\imout, \im) + \wptimse\mse(\imout, \im) + \wptireg\ptireg,
\end{equation}
where $\lpips$ is the perceptual loss~\cite{zhang2018perceptual}, $\mse$ denotes the MSE loss, $\wptimse$ denotes the weight for MSE the loss, $\ptireg$ is the locality regularization loss~\cite{roich2021pivotal}, and $\wptireg$ the weight for the regularizer.
The $\textrm{p}$ superscript denotes personalization, since we use $L2$ loss later for training our mapping networks as well.
Aside from the LPIPS and MSE loss, the locality regularization is enforced to make sure the PTI changes stay local without affecting other parts in the latent space.
Once, the network is fine-tuned, it can be seen as a personalized generator which now incorporates a prior based on the given subject, instead of the domain prior it originally has~\cite{nitzan2022mystyle}.

Finally, we can edit an inverted image by manipulating the latent code, which can be described as:
\begin{equation}
    \imout_\textrm{edit} = G(\wplus + \Delta \wplus;\theta_p),
\end{equation}
where $\Delta \wplus$ could be any editing directions from InterfaceGAN~\cite{shen2020interfacegan} or GANSpace~\cite{härkönen2020ganspace}.

In the next section, we discuss our proposed algorithm, starting by sampling frames from the input video, performing PTI on the selected frames to learn a personalized video prior, and then learning pose and expression mapper networks to represent different head pose and facial expressions on the personalized manifold.

\begin{figure}[htb!]
    \centering
    \includegraphics[width=1.0\textwidth]{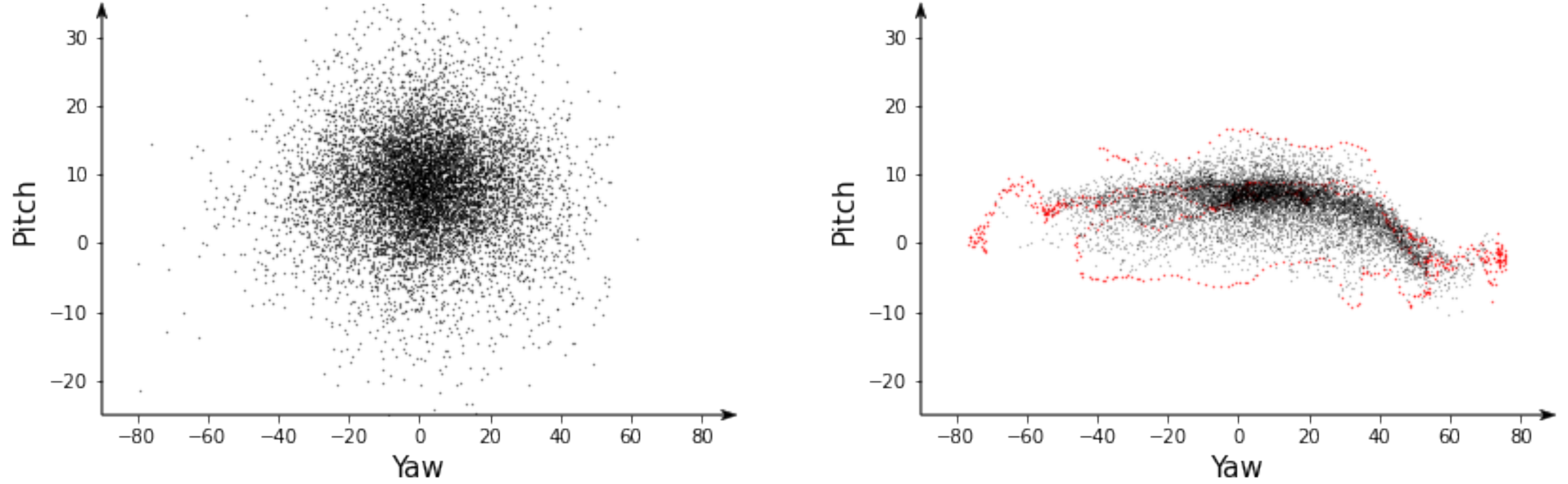}
    \caption{\boldstart{Head pose distribution of vanilla StyleGAN2 and personalized manifold.}
    We randomly synthesize 10k facial samples with the truncation parameter set to 0.5.
    Then we use DECA~\cite{DECA:Siggraph2021} to detect the yaw and pitch of the head.
    Most head poses have yaw inside $[\minus60, 60]$ and pitch in $[\minus20, 30]$, making it difficult to invert to head pose outside of this distribution.
    In the right figure, we randomly sample a linear combination of the pivots and synthesize novel views with StyleGAN.
    We highlight the head poses from the input video in red and the samples from the personalized manifold in black.
    The learned personalized manifold can reconstruct extreme yaw poses ($\ge60^\circ$) in the input video.
    }
    \label{fig:stylegan_poses}
\end{figure}

\section{Proposed Method}\label{sec:algorithm}

Given a monocular video of a dynamic portrait, we seek to produce an editable representation with fine controls over the pose, expression and appearance of the subject.
To achieve this goal, we develop an optimization pipeline with two stages: (a) learning the personalized video prior in StyleGAN, and (b) training pose and expression mapper networks.
To start, we discuss the first stage in Sec.~\ref{sec:manifold}.
Then, we describe the second stage, namely, how we train the pose and expression mappers, and how we use them to control different head poses and expressions in Sec.~\ref{sec:mapper}.
Last, we describe our loss design for the mappers in Sec.~\ref{sec:loss}.
Please refer to Fig.~\ref{fig:pipeline} for an overview of our optimization pipeline.


\subsection{Personalized Video Prior in StyleGAN}\label{sec:manifold}

Our main goal is to generate an edited portrait video with \edit{StyleGAN2}~\cite{Karras2019stylegan2}.
A naive way to address this is to perform GAN inversion on a frame-by-frame basis.
However, previous methods generally struggle with extreme viewing angles ($\sim90^\circ$).
This is because large head rotations ($>40^\circ$ in yaw, $>25^\circ$ in pitch) are less represented in the StyleGAN latent space, as shown in the left diagram in Fig.~\ref{fig:stylegan_poses}.
We notice that these out-of-domain (OOD) samples can be better represented with PTI~\cite{roich2021pivotal} as discussed in Sec.~\ref{sec:background}.
Another property we discovered is that interpolation between the pivots offers a smooth transition between different head poses (see the right diagram of Fig.~\ref{fig:stylegan_poses}), and we can represent the head motions as a linear combination of different pivot vectors, similar to the notion of linear blending in animation. 
As a result, our idea is to seek out these pivots in the given portrait video, and construct a manifold to allow animating the portrait by user control.

As discussed, when presented an image sequence, our task is to seek a personalized video prior $p$ that forms a local manifold $\Mp$ within the latent space of the fine-tuned StyleGAN $G(\cdot;\theta_p)$.
First, we preprocess the images by aligning and cropping the face following FFHQ~\cite{karras2019style}.
As the facial landmark detection could introduce slight inconsistencies and cause the results to flicker, we employ Gaussian filters to smooth out the noise~\cite{fox2021stylevideogan,tzaban2022stitch}.
Since there are many frames in a video, we seek to select the most useful frames as pivots to ensure efficient learning of the personalized manifold.
We design a sampling strategy to uniformly sample different head poses and expressions throughout the video to ensure good coverage of all possible facial changes.
To be more specific, we utilize an off-the-shelf face detector, DECA~\cite{DECA:Siggraph2021}, to estimate the yaw $\yaw$, pitch $\pitch$, neck pose $\neck$, jaw pose $\jaw$ and expression parameters $\expr$ of a given sequence.
We then stack the $\yaw$, $\pitch$ and $\expr$ channel-wise and perform K-means clustering~\cite{kmeans} to select the most prominent $K$ clusters.
While uniform sampling sometimes yields similar results, our sampling strategy would avoid oversampling cases where the pose or expression stay similar.
We find that this strategy provides good coverage of different expressions and head poses in the input video.
We define the $K$ samples from the video as $I = \{I_1, I_2, ..., I_K\}$.
Finally, we optimize the StyleGAN generator, similar to Eq.~\ref{eq:pti_obj}, with the following objective:
\begin{equation}
    \theta_p = \argmin_\theta \sum^K_{i=1} \frac{1}{K} \left[ \lpips(\imout_i, \im_i) + \wptimse\mse(\imout_i, \im_i) \right] + \wptireg\ptireg.
\end{equation}
The main difference from Eq.~\ref{eq:pti_obj} is that we are optimizing over the $K$ images instead of only one image.
From this point forward, we introduce a shorthand $\Gp = G(w^+; \theta_p)$.
Once we acquire the fine-tuned StyleGAN $\Gp$, we can define the $\beta$-dilated convex hull~\cite{nitzan2022mystyle} in the $\mathcal{W}+$ space as the local manifold $\Mp$:
\begin{equation}\label{eq:manifold}
    \Mp = \sum_{i=1}^K \alpha_i w_i, \ \ \ \ \ \ \ \text{  s.t.  } \sum_{i=1}^K \alpha_i = 1, \ \ \text{and } \alpha_i \ge \minus\beta,
\end{equation}
where $\alpha_i$ is the blending weight for pivot $w_i$.
As discussed in Sec.~\ref{sec:background}, interpolation between StyleGAN latent codes provides smooth transitions between different pivots.
We also notice that interpolating between poses at $90^\circ$ and $-90^\circ$ in $\Mp$ gives stable and high-quality renderings of the poses in between, while maintaining the same identity of the subject.

\boldstart{Discussion.}
Our proposed fine-tuning stage has two differences from MyStyle~\cite{nitzan2022mystyle}.
First, MyStyle uses 100 to 200 images of the given subject under different settings, including lighting, ages, expressions and hairstyles, whereas our method focuses on frames from the same input video where the subject can have similar appearance.
This difference make redundant samples and overfitting more likely to happen due to less variety in the training samples.
However, our proposed sampling strategy ensures the samples are different enough in poses or expressions by clustering and selecting the representative frames from the input video.
This way, our method can represent a wide variety of facial expressions and poses given a short video of around 20 seconds.
Another difference is that fine-tuning the StyleGAN without any regularizer $\ptireg$ could lead to degraded editability (see Fig.~\ref{fig:editing_reg}).
We theorize that this is because the fine-tuned StyleGAN overfits to the input video and causes other parts in the latent space to lose the smoothness of the original StyleGAN.
To prevent this, it is important to apply the locality constraint in our setting to keep the structure in the StyleGAN latent space relatively unchanged after the fine-tuning.
It is also beneficial to use StyleGAN's expressive latent space to synthesize expressions which are not observed in the input video.
We design a loss function that utilizes random expression objectives to encourage the network to synthesize these unseen expressions. 
More details are in Sec.~\ref{sec:mapper} and Sec.~\ref{sec:loss}.



\begin{figure}[t]
    \centering
    \includegraphics[width=1.0\textwidth]{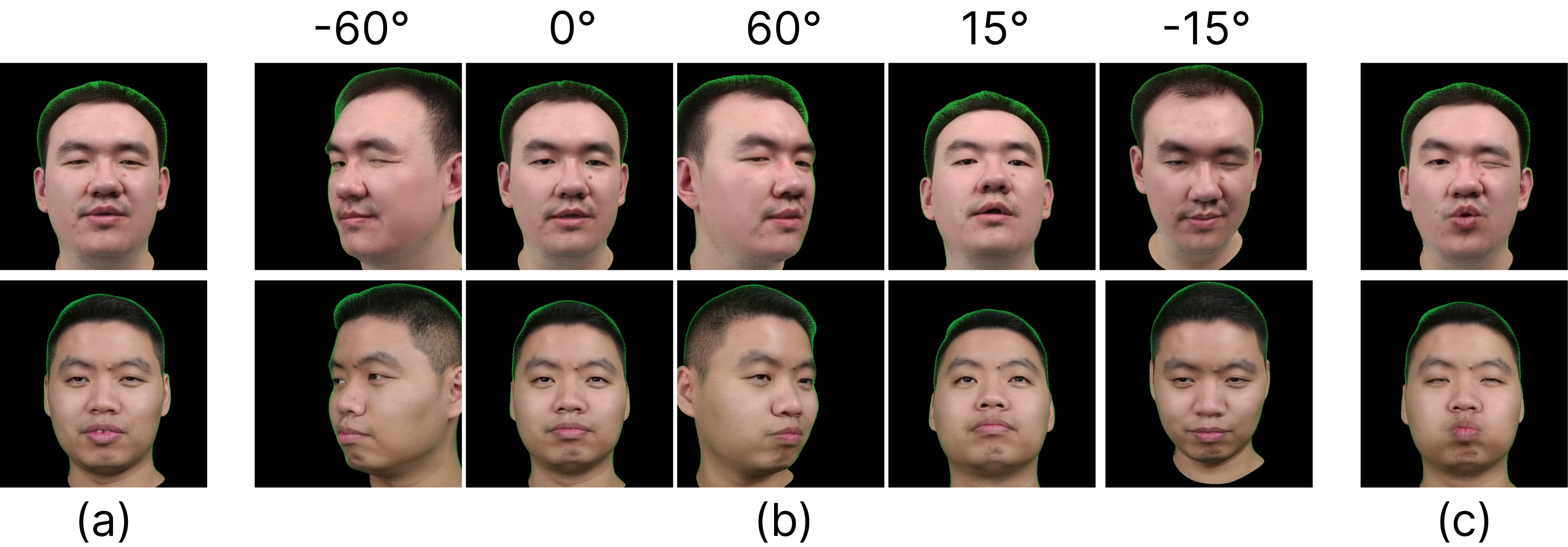}
    \caption{\boldstart{Examples of latent code shown in Sec.~\ref{sec:mapper}.}
    (a) The personalized manifold $\Mp$ provides a good representation of the subject, and it allows us to further edit the pose and expression.
    (b) The rotation code $\wrot$ enables free pose control of the synthesized portraits.
    Here we show specific yaw ($\minus60^\circ$ to $60^\circ$) and pitch ($\minus15^\circ$ to $15^\circ$) controls.
    (c) The expression code $\wexpr$ allows finer control over the facial expressions.
    }
    \label{fig:latent}
\end{figure}


\subsection{Pose and Expression Mapping Networks}\label{sec:mapper}

A key contribution of our proposed method is the ability to easily control the face with parameters like yaw, pitch and a 50-dimensional expression vector from DECA~\cite{DECA:Siggraph2021}.
To this end, we implement the pose and expression mappers, which are MLPs taking pose and expression coefficients as input and producing the corresponding latent code (see Fig.~\ref{fig:latent} for examples).


First, in order to change the head pose of the rendering, we could manipulate the latent code by moving in the personalized manifold $\Mp$.
Specifically, we employ an MLP $\rotmlp$ to calculate the latent code $\wrot$ as a linear combination of the pivots $\{w_i\}_{i=1}^K$ by
\begin{align}
    \wrot = \sum_{i=1}^K \rot_i w_i, \quad \text{and} \quad
    \{\rot_i\}_{i=1}^K = \rotmlp(\pitch, \yaw; \theta_\mathrm{r}),
\end{align}
where $\pitch$ denotes the pitch, $\yaw$ the yaw of the head and $\theta_\mathrm{r}$ represents the weights of $\rotmlp$.
We use the MLP to predict the blending weights instead of the latent code in $\mathcal{W}$ space or $\mathcal{W}+$ space, because the blending weights effectively constrain the latent code to be inside the manifold where all points are meaningful and would represent the subject (see the video results in supplementary materials for an example).
On the other hand, it is possible for editing directions in $\mathcal{W}$ and $\mathcal{W}+$ space to go beyond the personalized manifold and degrade the rendering significantly.

Lastly, we introduce expression control by learning a mapping network that takes as input the jaw pose $\jaw$ and the expression parameters $\expr$ and outputs the difference of the latent code $\Delta \wexpr$ in $\mathcal{W}+$ space.
Specifically, we have
\begin{equation}
    \Delta \wexpr = \exprmlp(\jaw, \expr;\theta_\textrm{e}).
\end{equation}
We then add it to the latent code $\wrot$:
\begin{equation}
    \wfinal = \wrot + \Delta \wexpr.
\end{equation}
In practice, we only predict the first 8 layers of the latent code instead of the full 18 layers.
This is because we would like to focus on geometry changes instead of other appearance changes.
We are effectively predicting a local change in the latent space to represent changes like lip motions during a talking sequence and facial expressions like raising of the eyebrows.
The local neighborhood of the latent space contains a rich and disentangled representation of the subject's possible facial expressions. 
\edit{Note that it is possible to move the latent code slightly outside the convex hull.
To regularize, we only use the first 8 layers of $\mathcal{W}+$ to reduce degrees-of-freedom, and we apply $\localloss$ (see Sec.~\ref{sec:loss}) to keep the changes small.}

Finally, we can then render the novel view $\imout$ with the fine-tuned StyleGAN generator $\Gp$:
\begin{equation}
    \imout = \Gp(\wfinal).
\end{equation}
Note that once the mapping network is trained, it can be run in real-time given the pose and expression parameters.

\subsection{Loss Design for Mapping Networks}\label{sec:loss}

Next, we fix the StyleGAN generator weight $\theta_p$, and train the mapping networks $\rotmlp$ and $\exprmlp$ with the loss design discussed in this section, 
optimizing for their weights $\theta_r$ and $\theta_e$.
As we aim to reconstruct an input video, we supervise the output with rendering losses like LPIPS~\cite{zhang2018perceptual} and $L2$ loss, which are denoted by $\lpips$ and $\pixel$, respectively.
Additionally, to ensure the synthesized identities are the same, we apply an identity loss $\idloss$~\cite{tov2021designing}, which uses a pretrained ArcFace~\cite{deng2019arcface} network to obtain identity features.

Since the input video would often show one combination of the head pose and expression, it is insufficient to train on the input video alone.
To avoid overfitting, we introduce regularization by synthesizing images with a slightly perturbed expression input.
Precisely, for each target view $\im$ and its corresponding jaw pose $\jaw$ and expression $\expr$, we render an additional rendering $I^{\textrm{e}}$ with $\Gp$ by changing the input to the perturbed parameters $\jaw' = \jaw + \epsilon$, $\expr' = \expr + \epsilon$, where we add a normal distribution $\epsilon \sim \mathcal{N}(0, \sigma)$.
Then we feed the new rendering $I^{\textrm{e}}$ to the encoder of DECA~\cite{DECA:Siggraph2021} to predict yaw $\yaw^e$, pitch $\pitch^e$, jaw pose $\jaw^e$, expression $\expr^e$, and neck pose $\neck^e$. and define the expression matching loss as:
\begin{equation}
    \exprloss = \| (\jaw^e, \expr^e) - (\jaw', \expr') \|^2_2.
\end{equation}
We found this loss to improve the disentanglement of expression and head pose, since the synthesized portrait has the same head pose but a slightly different facial expression.
This additional objective increases the diversity of facial expressions that the expression mapper can generate.
\edit{Visualization of the perturbed renderings is shown in Appendix~\ref{sec:perturb_renderings}.}
Furthermore, we enforce a pose consistency loss and an RGB consistency loss to make sure $I^{\textrm{e}}$ does not show large head motions.
We define the pose consistency loss as:
\begin{equation}
    \poseloss = \| \neck^e - \neck \|_{2}^{2},
\end{equation}
Similarly, $\neck$ is the neck pose for the input frame, and $\neck^e$ denotes the DECA-estimated neck pose of the perturbed rendering.
\edit{Note that we do not feed the neck pose $\neck$ as an input to the pose MLP.}
For the RGB consistency loss, we calculate a mask $m$ excluding the eyes and mouth regions, and then minimize the loss between the perturbed rendering $I^{\textrm{e}}$ and the original rendering $\imout$:
\begin{equation}
    \consistencyloss = \| m\odot (I^{\textrm{e}} - \imout) \|_2^2
\end{equation}

To further encourage the expression network to explore local regions in the manifold, we apply a regularization loss on the predicted latent $\Delta \wexpr$, namely,
\begin{equation}
    \localloss = \| \Delta \wexpr \|_2^2.
\end{equation}
Lastly, our training objective is defined as follows:
\begin{multline}
    (\theta_\textrm{r}, \theta_\textrm{e}) = \argmin_{\theta_\textrm{r}, \theta_\textrm{e}} \wlpips \lpips + \wpixel \pixel + \widloss \idloss \\
    + \wconsistency \consistencyloss + \wpose \poseloss + \wexprloss \exprloss + \wlocal \localloss,
\end{multline}
where we update the pose network weights $\theta_\textrm{r}$ and the expression network weights $\theta_\textrm{e}$.

\section{Experiments}

\begin{figure*}[hbtp]
    \centering
    \includegraphics[width=1.0\textwidth]{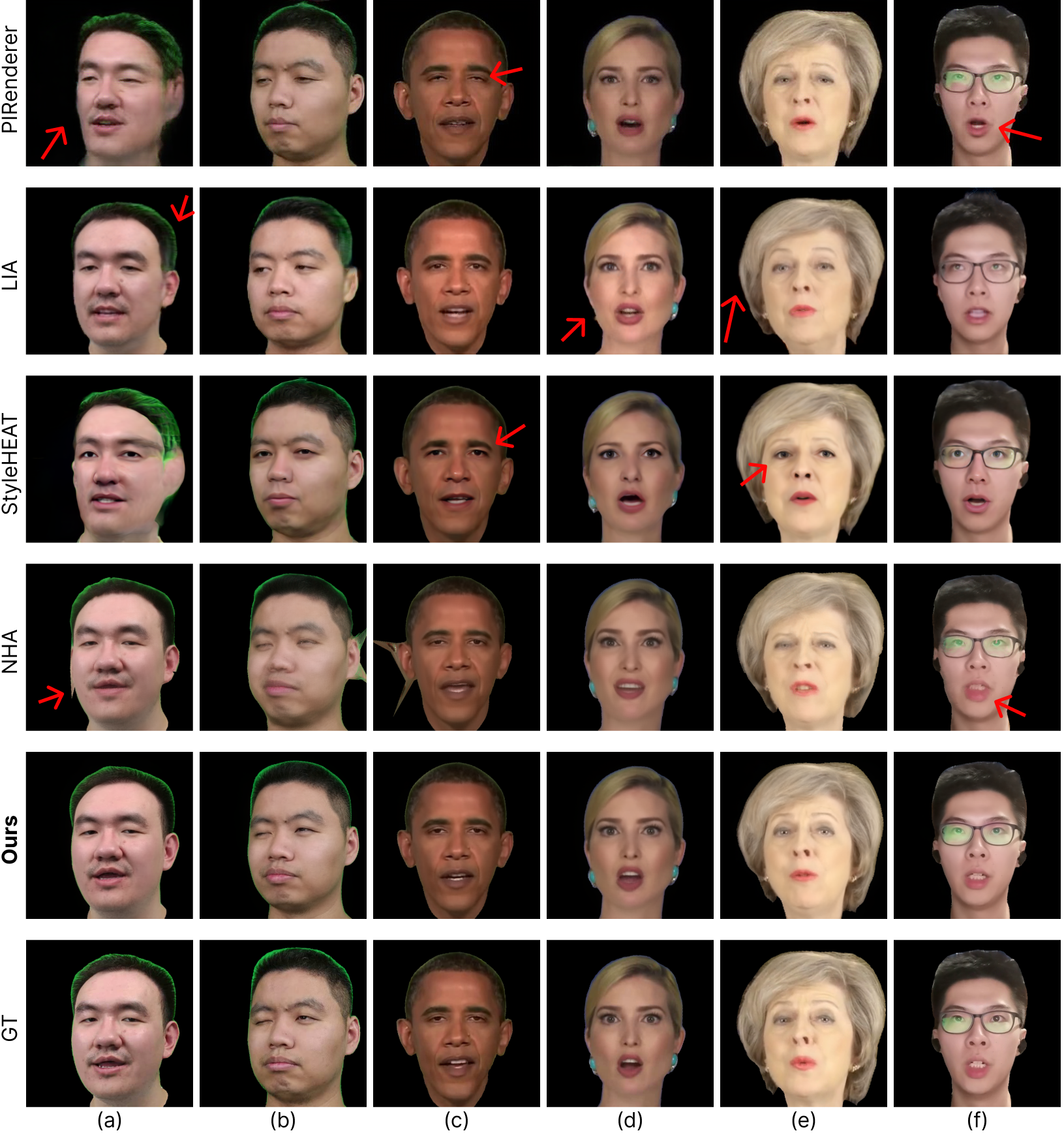}
    \caption{\boldstart{Visual results of different methods.}
    We show reconstruction results on the held-out views.
    Our method can predict better geometry and appearance of the subject.
    For example, in column (a), (b) and (c), we can see the ear reconstruction failed for NHA~\cite{grassal2021neural}.
    For 2D methods, they mostly fail to reconstruct the correct details and expression.
    To be more specific, PIRenderer~\cite{ren2021pirenderer} and StyleHEAT~\cite{styleheat} both fail for large viewpoint changes, with StyleHEAT generating overly-smoothed results, as shown by the red arrow in column (c).
    LIA~\cite{wang2022latent} overall has a tone shift and fails to reconstruct the details correctly, like the earring shown in (d).
    Our method offers faithful reconstruction of the head pose and expressions of the subject, as well as details, like the tint on the eyeglasses in (f).
    }
    \label{fig:qualitative}
\end{figure*}

\begin{figure}[hbt!]
    \centering
    \includegraphics[width=1.0\textwidth]{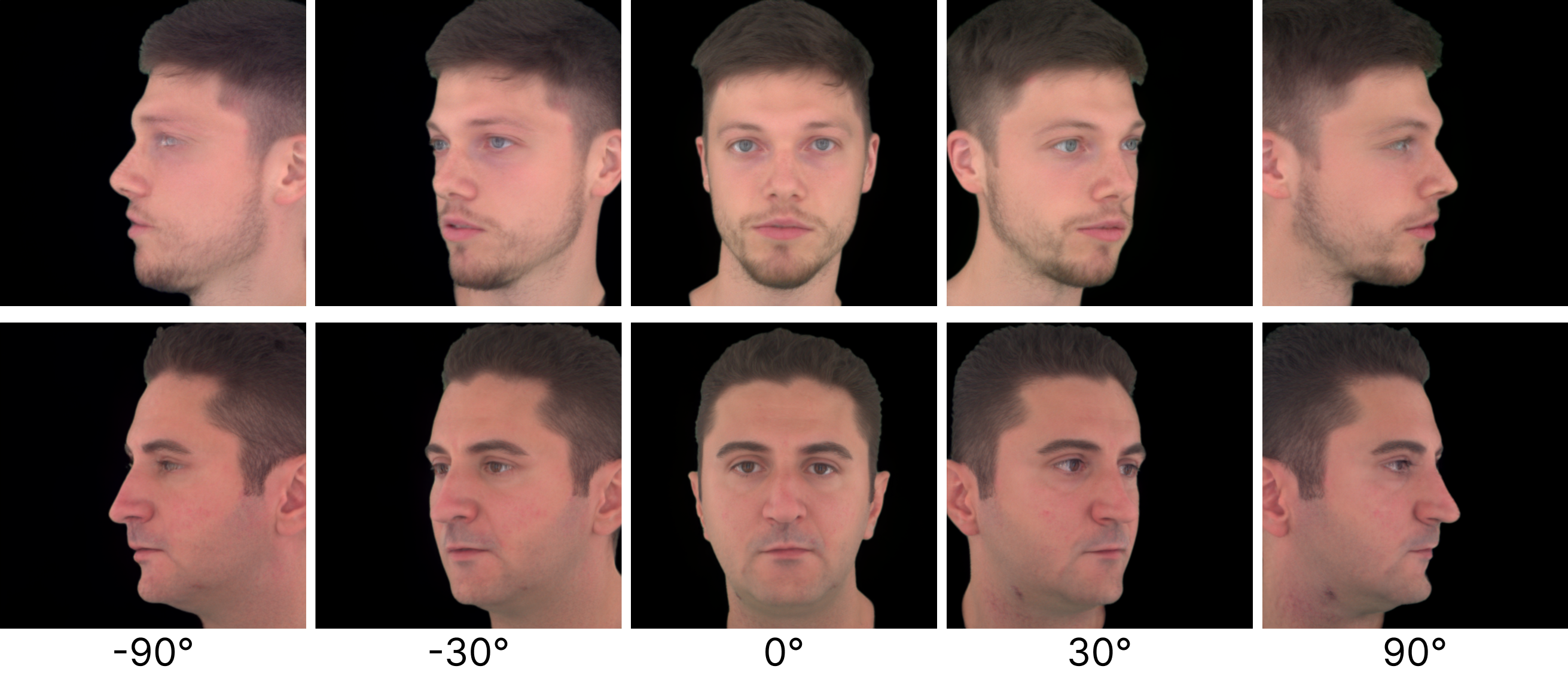}
    \caption{\boldstart{Visual results of our method with extreme head poses.}
    A key contribution of our method is to enable direct control over head poses with StyleGAN renderings.
    Our method can synthesize renderings with yaw from $\minus 90^\circ$ to $90^\circ$, which previous 2D-based methods cannot achieve.
    }
    \label{fig:rotation}
\end{figure}

We demonstrate the efficacy of our proposed method against SOTA 2D methods~\cite{ren2021pirenderer,wang2022latent,styleheat} and a 3D method~\cite{grassal2021neural} in this section.
We first describe implementation details in Sec.~\ref{sec:implementation}.
We show quantitative comparison in Sec.~\ref{sec:quanitative} and Table~\ref{tab:quantitative}.
Then we provide visual results in Sec.~\ref{sec:qualitative}, \crefrange{fig:qualitative}{fig:editing}.
Finally, in Sec.~\ref{sec:limitations}, we discuss limitations and possible future directions.

\subsection{Implementation Details}\label{sec:implementation}
We implement our algorithm using PyTorch~\cite{pytorch}.
We first run PTI~\cite{nitzan2022mystyle} with the LPIPS threshold set to 0.03, and max PTI steps set to 350, on the samples from the input video to acquire a personalized manifold.
Then we train our pose and expression mappers for 50k steps.
We set our learning rate for the MLPs to be $5\times10^{-4}$.
For the loss functions in Sec.~\ref{sec:loss}, we use $\wlpips = 10$, $\wpixel= 10$, $\widloss = 0.5$, $\wpose = 0.1$, $\wexpr = 0.1$, $\wconsistency = 1.0$, and $\wlocal = 0.5$.
We set $\sigma = 0.5$ for the perturbed expression parameters.
\edit{Further details can be found in Appendix~\ref{sec:app_impl}.}


For the dataset, we use the monocular video dataset from NHA~\cite{grassal2021neural} and NeRFBlendShape~\cite{Gao2022nerfblendshape}.
For the NHA dataset, we choose the first 750 frames as training data.
Due to the last few frames being corrupted in the original video data provided by the author, we choose frames 751 to 1450 as the evaluation dataset instead of frames 751 to 1500.  The
NeRFBlendShape dataset has videos with 3000 to 4000 frames, and we choose the last 500 frames as evaluation data.
\edit{We briefly describe dataset composition in Appendix~\ref{sec:dataset_comp}.}
Source code can be found on our website: \url{https://cseweb.ucsd.edu/~viscomp/projects/EGSR23PVP/}.


\begin{table}[t]
  {\caption{\boldstart{Evaluation of ours and baseline methods.}
  Our renderings offer the best visual quality across all metrics.
  2D-based methods struggle to handle viewpoints at larger angles, resulting in poor visual quality.
  While 3D-based methods like NHA can handle larger head motions, they fail to reconstruct good 3D geometry for in-the-wild videos, leading to inferior results across all metrics.
  }\label{tab:quantitative}}
  \begin{tabular}{l c c c}
    \toprule
    Methods       &   \textbf{PSNR}$\uparrow$ & \textbf{LPIPS}$\downarrow$ & \textbf{SSIM}$\uparrow$  \\
    \midrule
    PIRenderer    & 21.49 & 0.2431 & 0.7540  \\
    LIA           & 23.47 & 0.1796 & 0.8545 \\
    StyleHEAT     & 20.29 & 0.2494 & 0.7685 \\
    NHA           & 25.31 & 0.1538 & 0.8746 \\
    \textbf{Ours} & \textbf{26.63} & \textbf{0.1119} & \textbf{0.8803} \\
    \bottomrule
  \end{tabular}
\end{table}

\subsection{Quantitative Results}\label{sec:quanitative}
To evaluate the visual quality, we used metrics like peak signal-to-noise ratio (PSNR), structural similarity index (SSIM), and perceptual loss (LPIPS).
We evaluate all methods on the held-out views from the evaluation dataset discussed in Sec.~\ref{sec:implementation}.
As for the baseline methods, we evaluate the resolution at $512\times512$ and remove the background as a preprocessing stage to ensure fairness when compared to methods like NHA~\cite{grassal2021neural}.
We also set the first frame of the video as the source frame for 2D-based methods and train NHA on the same training set as our method.
We show the quantitative results in Table~\ref{tab:quantitative}.
Our method offers the best performance among all the methods across all metrics.
Specifically, it can handle difficult head poses like $90^\circ$ to the left and to the right.
Moreover, the personalized manifold enhances the details in the reconstructed images as the StyleGAN generator is fine-tuned to produce highly-similar images as the input video.
Note that although our method is a 2D-based method, it still can produce multi-view consistent imagery.
This is because the head rotations are mostly represented by interior points of the personalized manifold.
And we observe that the interpolation between the pivots show good consistency for the identity and the geometry.
As a result, our method can provide better visual quality than 3D methods, such as NHA.
Finally, our method supports real-time rendering thanks to its lightweight mapping network.
The inference time for our method is 0.018s, which is about 54 FPS, on an RTX 3080 GPU.

\begin{figure}[hbt!]
    \centering
    \includegraphics[width=1.0\textwidth]{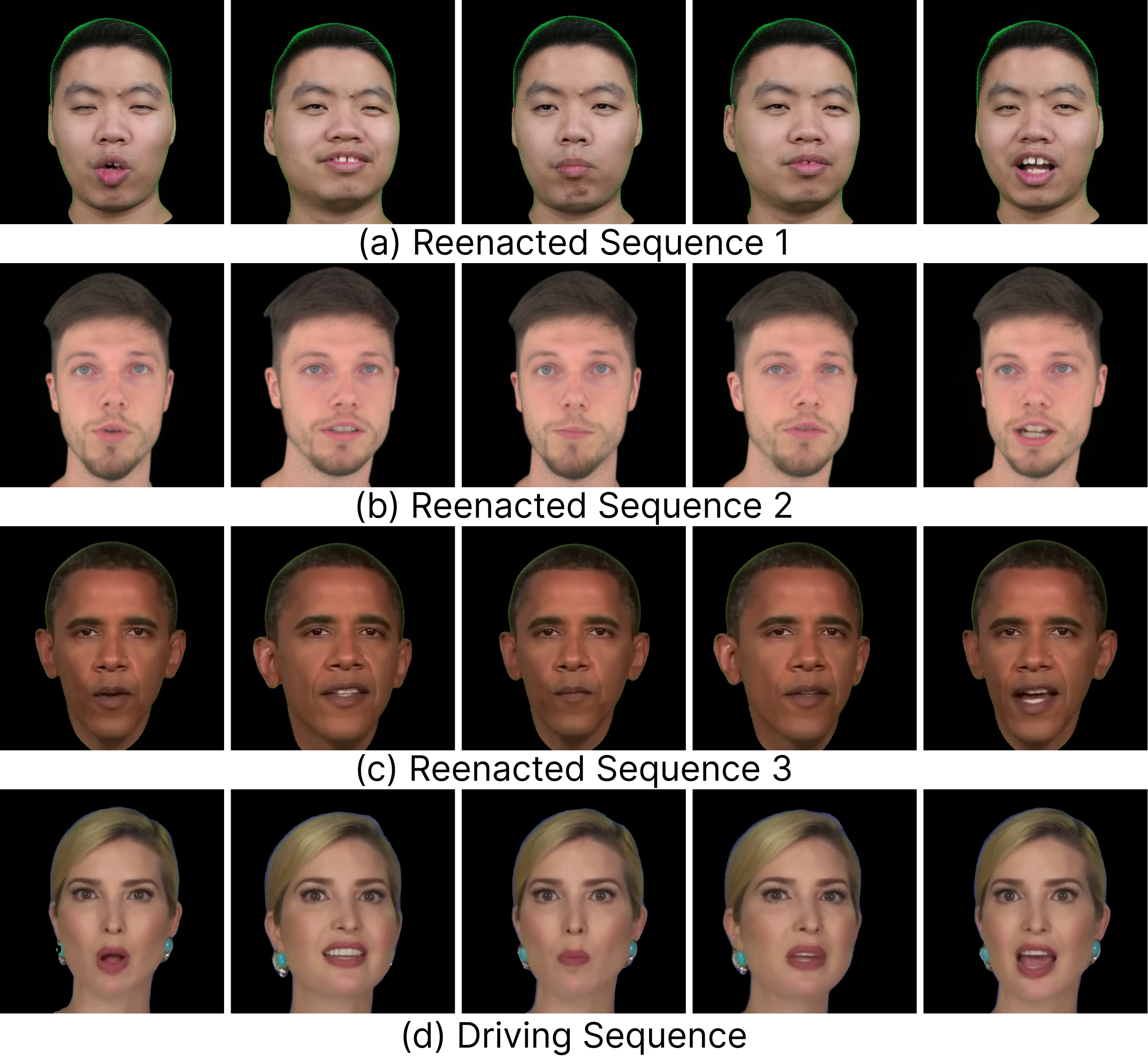}
    \caption{\boldstart{Reenactment results of our method.}
    We show reenacted sequences in (a), (b) and (c), and the driving sequence in (d).
    Our method is able to transfer the expressions across different subjects.
    }
    \label{fig:reenact}
\end{figure}

\begin{figure*}[hbt!]
    \centering
    \includegraphics[width=0.98\textwidth]{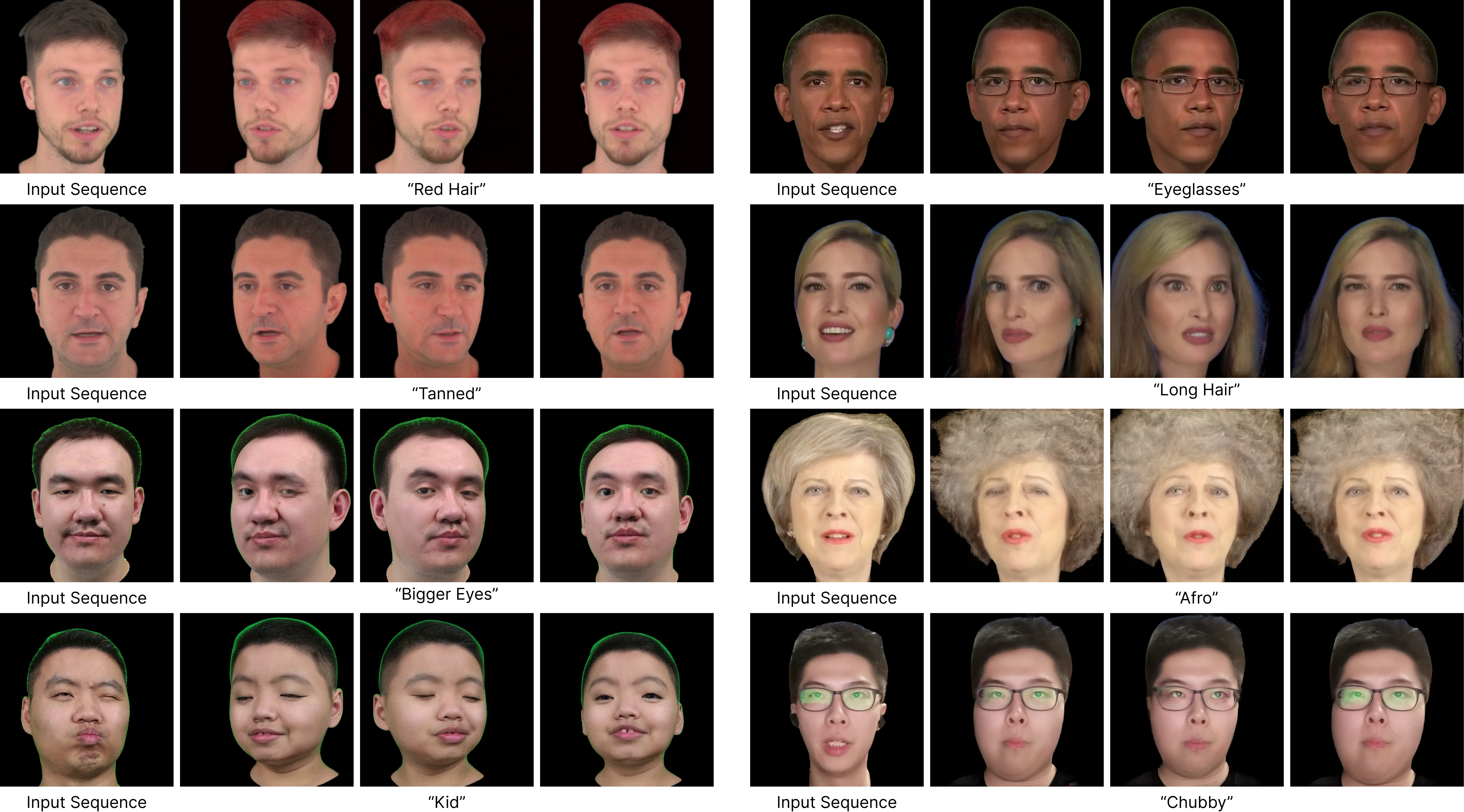}
    \caption{\boldstart{Editing results of our method.}
    We show a random frame from the input video sequence and the edited renderings with different head pose and facial expressions.
    Below each inset, we show the StyleCLIP prompt to edit the latent code and produce different editing results.
    Our method preserves the same identity and edits across different expression and pose changes.
    For example, in the "Kid" prompt, we show various expressions of the subject as a younger avatar and keep similar identities.
    Also, note the tint in the "Chubby" prompt is preserved and shows view-dependent changes across different viewpoints.
    }
    \label{fig:editing}
\end{figure*}

\begin{figure*}[hbt!]
    \centering
    \includegraphics[width=1.0\textwidth]{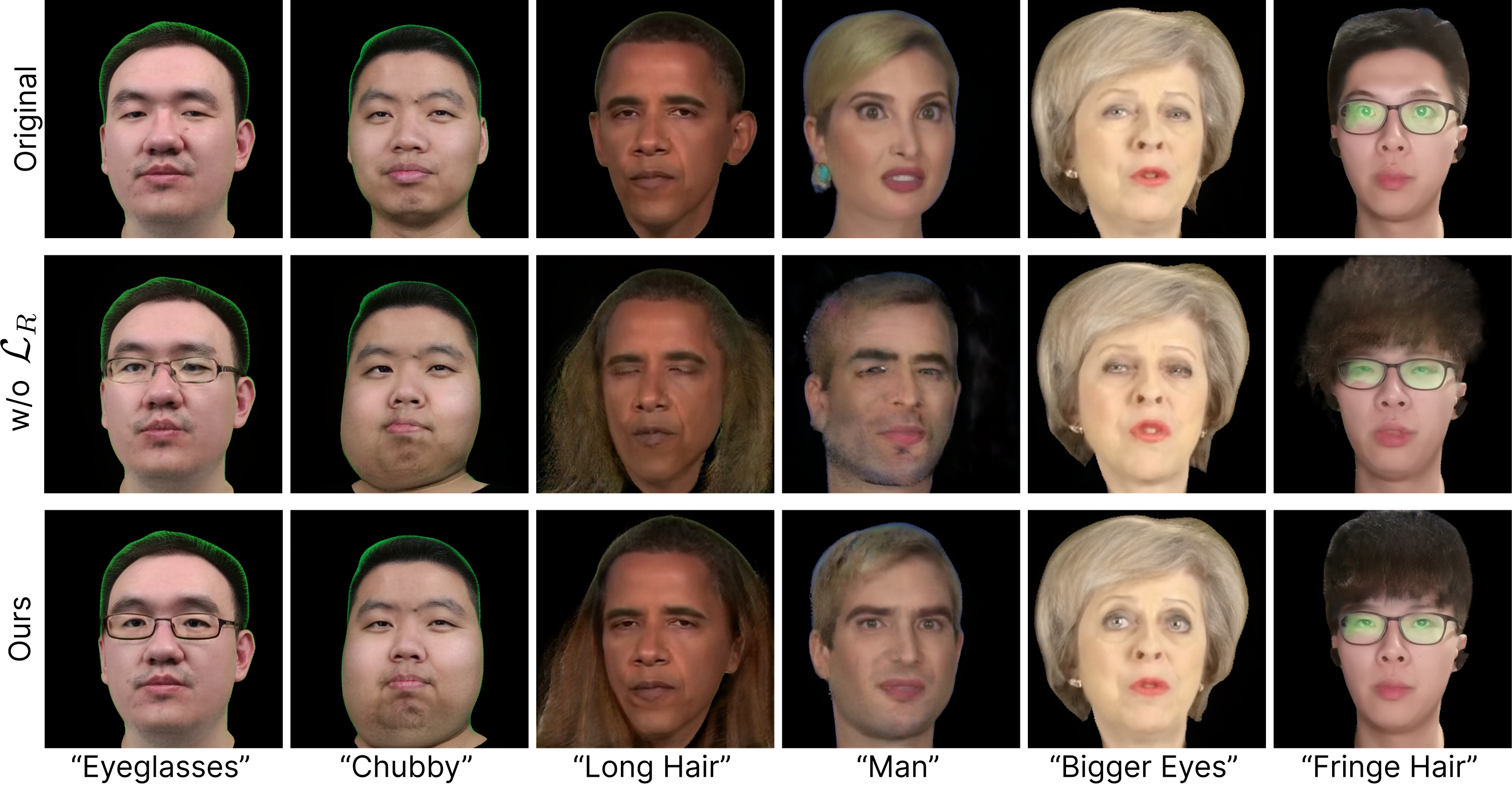}
    \caption{\boldstart{Effects of the regularizer on the editing results.}
    We observe that using selected frames from the input video could make it easier to overfit.
    Consequently, different from MyStyle~\cite{nitzan2022mystyle}, we apply a regularizer during the fine-tuning stage to maintain the editing ability of StyleGAN.
    }
    \label{fig:editing_reg}
\end{figure*}

\subsection{Qualitative Results}\label{sec:qualitative}
Figure~\ref{fig:qualitative} demonstrates the visual results of our method against other baseline methods.
Our algorithm is able to reconstruct face poses at extreme viewpoints, while previous methods like StyleHEAT and PIRender fail to generate reasonable results.
We can observe distorted face cheeks in (a) for both methods and even some repeating patterns near the back of the head.
Moreover, StyleHEAT often produces changes in the subject's eyes, as can be seen in (b) and (c).
The StyleHEAT results are overly smoothed, leading to missing details like wrinkles seen in (d) and (e).
On the other hand, PiRenderer, while keeping some details, misses the mouth texture, such as missing teeth shown in (d) and (f). 
As for LIA, it produces tone shifts in all the results, as well as missing details, for example the earrings in (d) and the tint on the eyeglasses in (f). 
LIA also fails to reconstruct the expression correctly, and it can be observed in (b), where the subject's eyes are fully open instead of half open as shown in the GT.
NHA has a hard time reconstructing faces captured with fewer head poses.
As shown in (a), (b) and (c), we can observe that the ears are heavily distorted due to NHA's optimization process.
Moreover, it fails to reconstruct the teeth region in (f) and it does not match the expression in (d) and (e).
\edit{We show additional comparison with NeRFBlendShape in Appendix~\ref{sec:nbs_comp}.}

In Figs.~\ref{fig:latent} and \ref{fig:rotation}, we demonstrate how our method can be used to generate rotated views of the subject, while fixing the expression.
Previous methods~\cite{styleheat} often have a difficult time handling extreme viewpoints as these viewpoints are out of the distribution of the pretrained StyleGAN generator.
Since we construct a personalized manifold, our method can represent extreme head poses well, as long as they are presented in the input views.
Note that we choose to learn the blending weights in the manifold, instead of an editing direction which controls the head motion.
This design choice is because we found that while the linear interpolation can be smoothly changing between pivots, it does not fully represent the desired head motion.
In other words, the personalized manifold might still contain some nonlinearity which causes the actual rotation to follow along a curved path, instead of a linear path.
The learned MLP ensures that we can follow a correct trajectory to represent different head poses in a high-dimensional latent space.

Moreover, we show reenactment results in Fig.~\ref{fig:reenact}.
Our method can be used to reenact different subjects with the provided expression parameters.
We extract the FLAME parameters $\yaw$, $\pitch$, $\jaw$ and $\expr$ from the driving sequence in (d) and feed it to the learned mapping networks for each identity in (a), (b) and (c).
Then, we can produce reenactment results with good accuracy.
For instance, in the rightmost column, we show reenactment results of all avatars with the mouth open expression.
Each avatar can do this expression properly with their unique and accurate teeth texture, which is often difficult to recover in previous methods.
We notice that since the FLAME parameters might have different distributions for each individual, it is better to renormalize the input parameters with the mean and standard deviation of the source and target.

Finally, we show editing results in Fig.~\ref{fig:editing}.
In the figure, we demonstrate various editing tasks using StyleCLIP~\cite{patashnik2021styleclip}.
Our method can be adapted to use other editing methods like InterfaceGAN~\cite{shen2020interfacegan} and GANSpace~\cite{härkönen2020ganspace}.
Note that our method provides consistent renderings and maintains the edits even after the head is rotated.
Furthermore, we show that the edited avatar can still produce various expressions, as shown in the ``Kid'' inset.
The edited latent code also shows good multi-view consistency.
Most notably, the ``Eyeglasses'' inset shows rotated versions of the subject with the eyewear with good geometry across different head poses.
Some view-dependent effects are retained, for instance, the tint on the eyeglasses in the ``Chubby'' inset.

It is worth noting that while we fine-tune the StyleGAN generator on selected video frames, the editing capability is not impacted thanks to the regularizer $\ptireg$.
Comparison of the regularizer can be found in Fig.~\ref{fig:editing_reg}.
Since we only have a handful of views of the given subject, it is highly likely for the fine-tuning stage to overfit to the distribution of these views.
Therefore, it is critical to enforce the regularizer to avoid degraded results in editing.
For instance, without the regularizer, the edited images often show color shift artifacts as shown in ``Eyeglasses'' and ``Long Hair''.
Additionally, the details and expressions are not retained after the edits, as shown in the ``Chubby'' and ``Man'' insets.

\subsection{Limitations and Future Work}\label{sec:limitations}

While our proposed method shows promising results in enabling editing of pose, expression, and appearance of digital avatars, there are several limitations that we would like to address in future work.

First, our method only handles the regions within the face alignment bounding box, excluding the back of the head and upper body.
Future work could focus on improving the StyleGAN generator or cropping to incorporate these regions, allowing for a more comprehensive representation of the subject.

In addition, our approach requires learning a personalized manifold for each subject, which can take some time for optimization.
To be more specific, learning the manifold takes around 3 hours for 200 pivot images on an RTX 3080 and training the mapping networks takes around 4 hours on an NVIDIA A10 24 GB GPU.
Note that after the optimization stage, our pipeline runs in real-time at 54 FPS on an RTX 3080.
It could be interesting to explore meta-learning and have a network that outputs a personalized manifold directly by looking at images.
Also, the mapping network could be pretrained on a large-scale dataset and apply to each avatar directly or with a fast fine-tuning stage.
This could potentially remove the optimization overhead and enable faster adaptation to new subjects.

We also note that sometimes the gaze or eye regions are not perfect.
However, this is not bound by our network design, but rather the performance of the DECA algorithm.
For future work, it is possible to swap out DECA for better facial expression detectors like EMOCA~\cite{danecek2022emoca} and add some gaze regularization.

Finally, our method works best for interpolation, but may not perform well when extrapolating beyond the training data.
Future work could explore additional regularization methods or data augmentation techniques to enable extrapolation and improve the overall generalization of the model.

\subsection{Ethical considerations}\label{sec:ethics}
The success of recent approaches in synthesizing photorealistic editable representations of a given subject, as achieved in this paper, has necessitated the introduction of various methods to detect if an image is fake~\cite{masood2022deepfakes}. However, the best methods can often be used as critics in the training paradigm of the state-of-the-art generative or editing models, in order to avoid detection. Additionally, as models become better, existing detection methods may be unable to scale. To prevent the misuse of editing methods, the development of more robust detection and verification techniques is paramount. 
\section{Conclusions}

We propose a novel algorithm that encodes a monocular portrait video into a personalized manifold to enable editing on pose, expression, and appearance.
Our approach selects useful pivots from the video sequence, allowing for efficient learning of the personalized manifold.
We also design loss functions to learn pose and expression mapping networks, enabling granular control of the rendering given only a single video.
Moreover, our method provides good editing capability through various StyleGAN editing methods.
Overall, our work significantly contributes to the development of digital avatars by making them more interactive, engaging, and personalized.
\section*{Acknowledgement}
This work was funded in part by a Qualcomm FMA Fellowship, ONR grants N000142012529, N000142312526, an NSF Graduate Fellowship, NSF grants 2100237 and 2120019, and the Ronald L. Graham Chair.
We also acknowledge gifts from Adobe, Google, Meta and a Sony Research Award.
All of the datasets mentioned in this paper were solely downloaded and used by University of California San Diego.

{\small
\bibliographystyle{ieee_fullname}
\bibliography{egbib}
}

\section*{Appendix}
\appendix

\section{Perturbed Renderings for Expression Matching Loss}\label{sec:perturb_renderings}

\begin{figure}[htb!]
    \centering
    \includegraphics[width=\textwidth]{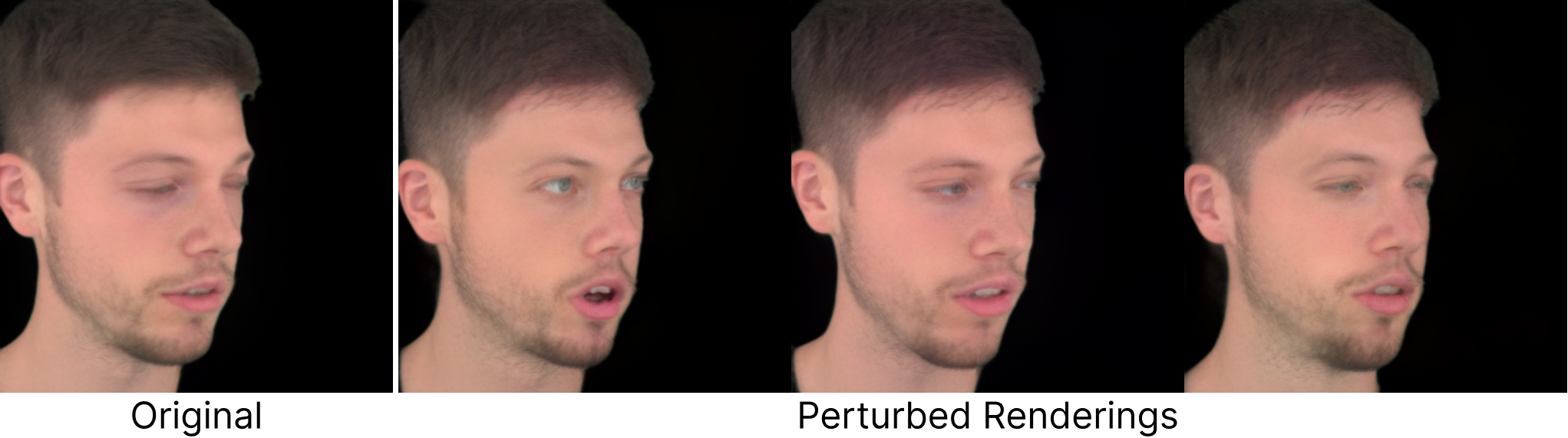}
    \caption{\boldstart{Renderings from the perturbed parameters $\gamma'_i$.}}
    \label{fig:perturbed}
\end{figure}

We show the renderings from the perturbed parameters $\gamma'_i$ in Fig.~\ref{fig:perturbed}.
The expression matching loss $\exprloss$ helps the network learn unseen expressions through perturbing the parameters and matching the predictions from DECA.

\begin{figure}[htb!]
    \centering
    \includegraphics[width=\textwidth]{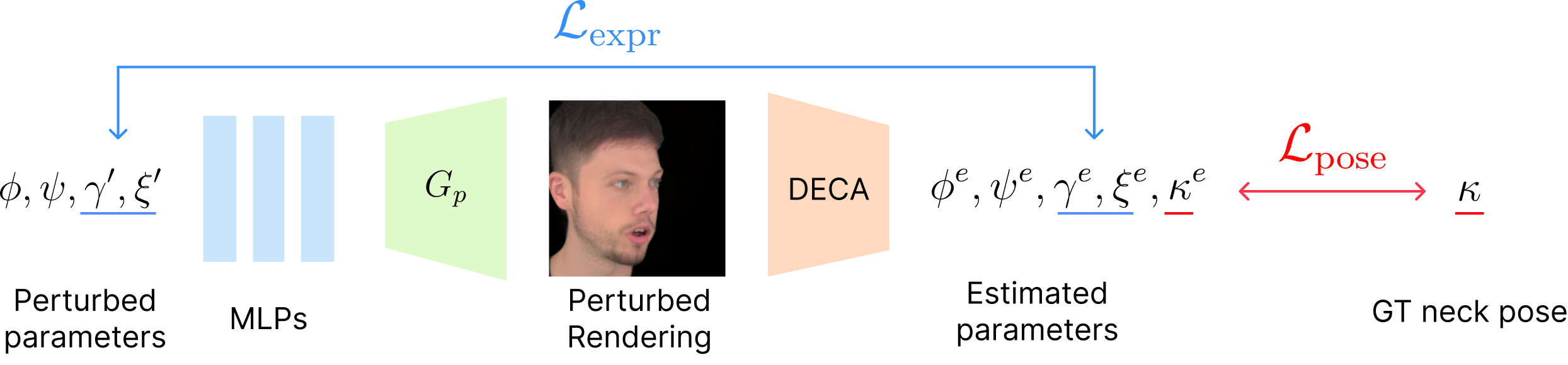}
    \caption{\boldstart{Illustration of our supervision.}}
    \label{fig:supervision}
\end{figure}

\section{Implementation Details}\label{sec:app_impl}

We discuss the implementation details of our algorithm in this section.
$\rotmlp$ is a 2-layer MLP with leaky ReLU as the nonlinear activation layer in the middle and Tanh as the output layer.
There are 128 channels for the hidden layer.
Instead of outputting the blending weights directly, the network predicts the residual weights from the centroid, which has equal weights from all pivots, to ensure the initialization is around the centroid (since network output is small initially).
Then it linearly combines the blending weights into a latent code in the $\mathcal{W}+$ space. 
Additionally, to prevent the MLP from generating values lower than negative beta, we shift the output values by beta and apply a SoftPlus function.
This would effectively bound the values.
$\exprmlp$ is composed of the same 2-layer MLPs, except that there are 8 of them to control the first 8 layers of the latent code in the $\mathcal{W}+$ space.
The predicted values are added to the latent code output from $\rotmlp$.

Another implementation detail is that we perform a normalization stage for reenactment.
The idea is that the distributions of expressions are different for each video.
Since we only train on a single input video, it is possible to overfit to the given sequence.
As a result, we normalize the driving features with standard deviation and mean of the expressions from the source and driving video before input it to the $\exprmlp$ for reenactment.

For the clustering stage discussed in Sec.~4.1, we use K=200 in our experiments.
It is important when the expression distribution in the data is uneven (e.g. same expressions for a long time) as the clustering can select more representative frames.

In terms of rendering resolution, we downsample our results from $1024\times1024$ to $512\times512$ for fair comparison with other baseline methods.
However, our method can run in $1024\times1024$ resolution with similar efficiency (removing the downsampling stage), given input videos with $1024\times1024$ resolution.

\begin{figure*}[htb!]
    \centering
    \includegraphics[width=0.9\textwidth]{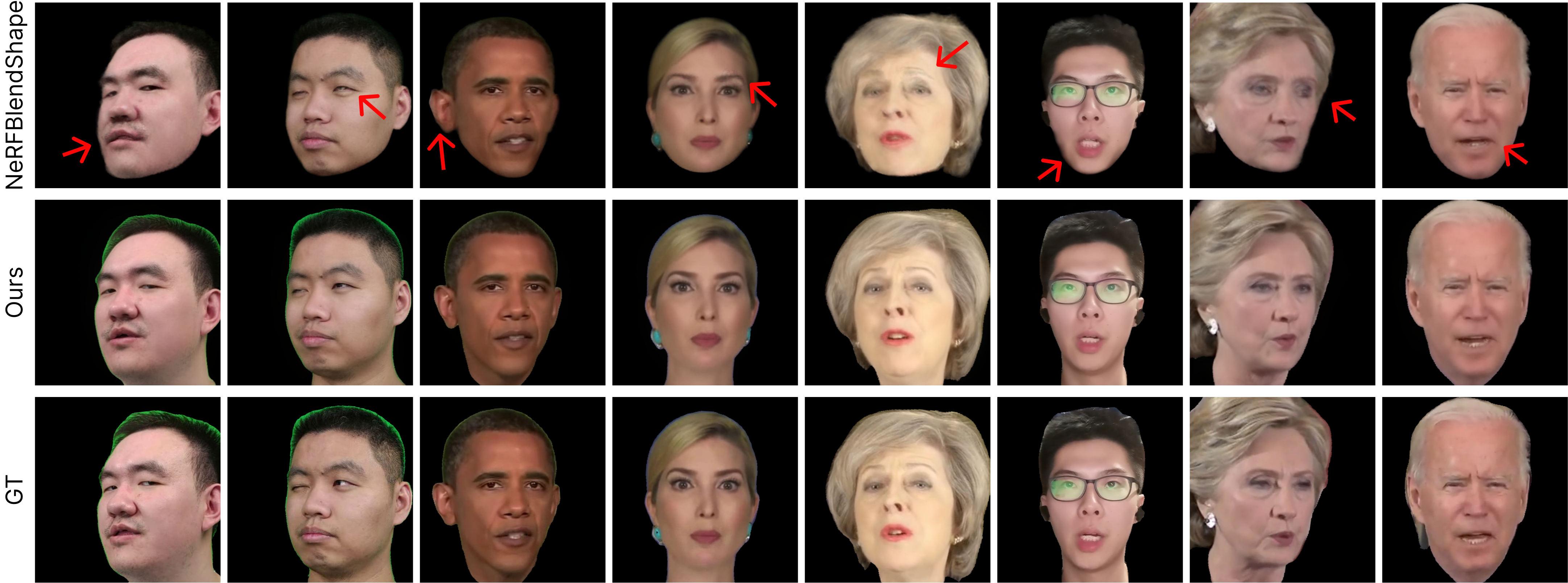}
    \caption{\boldstart{Qualitative comparison with NeRFBlendShape}
    Our method provides sharper details compared to NeRFBlendShape~\cite{Gao2022nerfblendshape}.
    While NeRFBlendShape shows more accurate geometry, it produces artifacts when transitioning between different expressions, as highlighted above.
    Please zoom in to better see the details.
    }
    \label{fig:nbs_qual}
\end{figure*}

\section{Dataset Composition}\label{sec:dataset_comp}

We show the range of head poses in Table~\ref{tab:dataset_comp}.
Dataset names starting with ``id'' denote the subjects from the NeRFBlendShape dataset, whereas names starting with ``person'' denote subjects from the NHA dataset.
It is difficult to determine the number of expressions as they are often transitioning from one to another (e.g. from smiling to grinning).
However, in the NHA and NeRFBlendShape dataset, there are different expressions like smiling, winking and puffing.

\begin{table}[htb!]
{\caption{\boldstart{Composition of our dataset.}
We show the range of head poses (in degrees) of each video sequence.
Our evaluation dataset contains a wide range of head poses.
}\label{tab:dataset_comp}}
\centering
\resizebox{0.9\textwidth}{!}{%
\begin{tabular}{lrrrr}
\toprule
Dataset & \multicolumn{1}{l}{Min Yaw} & \multicolumn{1}{l}{Max Yaw} & \multicolumn{1}{l}{Min Pitch} & \multicolumn{1}{l}{Max Pitch} \\
 \midrule
id1        & -45.36 & 32.29 & -23.25 & 17.06 \\
id2        & -57.74 & 35.85 & -39.35 & 15.70 \\
id3        & -13.43 & 14.38 & -20.60 & 12.44 \\
id4        & -29.09 & 12.39 & -14.98 & 14.95 \\
id5        & -17.22 & 12.27 & -5.42  & 11.41 \\
id6        & -9.62  & 2.31  & -9.67  & -1.38 \\
id7        & -39.96 & 36.23 & -22.38 & 13.24 \\
id8        & -4.98  & 10.09 & -20.16 & -4.51 \\
person0000 & -75.56 & 75.92 & -35.00 & 10.49 \\
person0004 & -70.62 & 72.65 & -37.2  & 17.85 \\
\bottomrule
\end{tabular}
}
\end{table}

\section{Additional Comparison with NeRFBlendShape}\label{sec:nbs_comp}

We show additional experiments with NeRFBlendShape~\cite{Gao2022nerfblendshape} in Table~\ref{tab:nbs_quantitative} and Fig.~\ref{fig:nbs_qual}.
In Table~\ref{tab:nbs_quantitative}, we provide two different numbers: original and masked.
The original set is the same as Table~\ref{tab:quantitative}, where we evaluate the whole image.
The masked set uses the alpha values from NeRFBlendShape as the mask and only evaluate the face regions.
While our method is purely 2D, it shows comparable performance to SOTA 3D methods.
In Fig.~\ref{fig:nbs_qual}, we demonstrate that our method produces sharper details, whereas NeRFBlendShape could produce ghosting artifacts.

\begin{table}[htbp!]
  {\caption{\boldstart{Evaluation of ours and NeRFBlendShape.}
  We show quantitative results from the NeRFBlendShape data.
  Our method demonstrates comparable performance.
  Moreover, our method offers better and sharper image in terms of the LPIPS.
  }\label{tab:nbs_quantitative}}
  \resizebox{0.8\textwidth}{!}{%
  \begin{tabular}{l c c c c}
    \toprule
    Methods        &  Masked & \textbf{LPIPS}$\downarrow$ & \textbf{PSNR}$\uparrow$ & \textbf{SSIM}$\uparrow$  \\
    \midrule
    NeRFBlendShape & \xmark & 0.1645 & 20.20 & 0.8689  \\
    \textbf{Ours}  & \xmark & \textbf{0.1119} & \textbf{26.63} & \textbf{0.8803} \\
    \midrule
    NeRFBlendShape & \cmark & 0.0981 & \textbf{32.05} & \textbf{0.9323}  \\
    \textbf{Ours}  & \cmark & \textbf{0.0856} & 31.76 & 0.9256 \\
    \bottomrule
  \end{tabular}
}
\end{table}

\end{document}